\documentclass[10pt,twocolumn,letterpaper]{article}

\usepackage{cvpr}              
\definecolor{cvprblue}{rgb}{0.21,0.49,0.74}
\usepackage[pagebackref,breaklinks,colorlinks,allcolors=cvprblue]{hyperref}
\usepackage{booktabs}    
\usepackage{colortbl}    
\usepackage{xcolor}      
\usepackage{graphicx} 
\usepackage{mathtools}
\usepackage{xcolor}   
\usepackage{pifont}   
\usepackage{multirow}

\def\paperID{9346} 
\def\confName{CVPR}
\def\confYear{2026}

\title{Personalized Reward Modeling for Text-to-Image Generation}

\author{
Jeongeun Lee \\
Yonsei University \\
{\tt\small ljeadec31@yonsei.ac.kr}
\and
Ryang Heo \\
Yonsei University \\
{\tt\small ryang1119@yonsei.ac.kr}
\and
Dongha Lee\thanks{Corresponding author} \\
Yonsei University \\
{\tt\small donalee@yonsei.ac.kr}
}
\newcommand{\proposed}{{PIGReward}\xspace}
\newcommand{\bench}{{PIGBench}\xspace}

\newcommand{\smallsection}[1]{{\vspace{0.05in} \noindent \bf {#1.\hspace{5pt}}}}
\definecolor{DarkGreen}{RGB}{30,130,30}
\newcommand{\cmark}{\textcolor{DarkGreen}{\ding{51}}} 
\newcommand{\xmark}{\textcolor{red}{\ding{55}}}       

\begin{document}
\maketitle

\begin{abstract}
Recent text-to-image (T2I) models generate semantically coherent images from textual prompts, yet evaluating how well they align with individual user preferences remains an open challenge.
Conventional evaluation methods, general reward functions or similarity-based metrics, fail to capture the diversity and complexity of personal visual tastes.
In this work, we present \proposed, a personalized reward model that dynamically generates user-conditioned evaluation dimensions and assesses images through CoT reasoning.
To address the scarcity of user data, \proposed\ adopt a self-bootstrapping strategy that reasons over limited reference data to construct rich user contexts, enabling personalization without user-specific training.
Beyond evaluation, \proposed\ provides personalized feedback that drives user-specific prompt optimization, improving alignment between generated images and individual intent.
We further introduce \bench, a per-user preference benchmark capturing diverse visual interpretations of shared prompts.
Extensive experiments demonstrate that \proposed\ surpasses existing methods in both accuracy and interpretability, establishing a scalable and reasoning-based foundation for personalized T2I evaluation and optimization.
Taken together, our findings highlight \proposed\ as a robust step toward individually aligned T2I generation.
\end{abstract}
\section{Introduction}

Recent text-to-image (T2I) generative models~\cite{ramesh2022hierarchical, saharia2022photorealistic, podell2023sdxl, pernias2023wurstchen} have shown remarkable performance in generating high-quality images from natural language prompts.
Moving beyond generating generic, one-size-fits-all images, following works explores \textit{personalization} in T2I generation to better align with individual preferences~\cite{xu2025personalized, wei2025personalized, dunlop2025personalized}.
These methods typically adopt a history-driven approach~\cite{chen2024tailored,dang2025personalized,kim2025draw}, leveraging user-specific references (e.g., prior image preferences or interactions) to reflect unique visual tastes.

\begin{figure}[t]
    \centering
    \includegraphics[width=\linewidth]{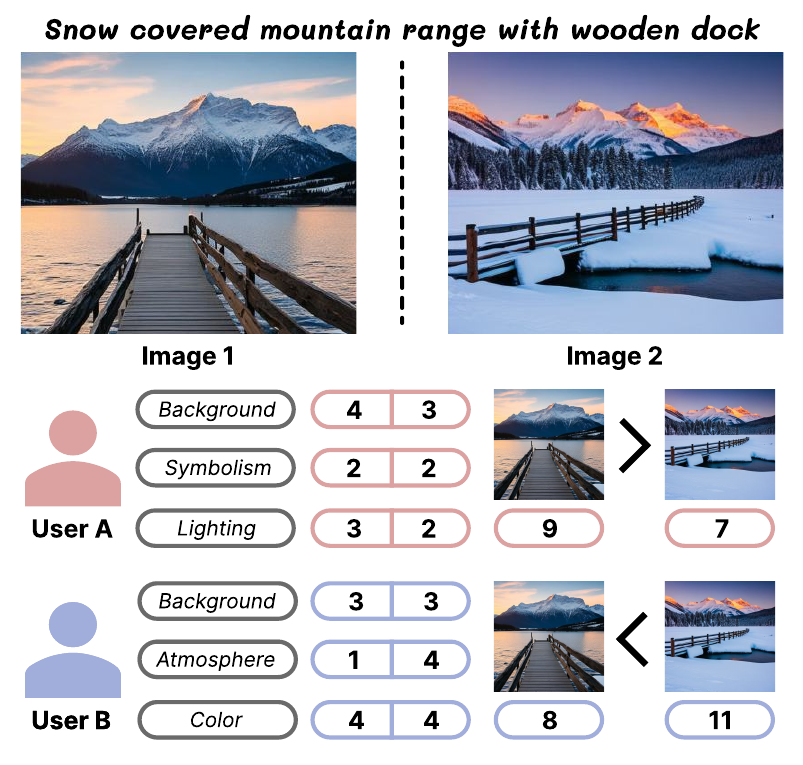}
    \vspace{-0.6cm}
    \caption{Each image in the pair is evaluated across multiple dimensions per user. Since two users prioritize different criteria, their overall judgments of which image is better differs.}
    \label{fig:intro}
\end{figure}

However, evaluating personalization remains a significant challenge under conventional T2I evaluation methods. 
Most widely-used approaches are based on reward functions fine-tuned on large-scale general human preference datasets~\cite{kirstain2023pick, wu2023human, xu2023imagereward}.
They reduce human preferences to a single scalar score, failing to consider their inherent complexity.
This motivates multi-dimensional evaluation~\cite{xu2024visionreward, zhang2024learning, liang2024rich}, which introduces a small set of pre-defined dimensions (e.g., Alignment, Detail).
More recently, the strong reasoning capabilities of large vision–language models (LVLMs) have inspired their use as generative reward models~\cite{zhou2025multimodal, wang2025unified, wang2025unifiedcot}, offering interpretable explanations.
Yet they still rely on fixed criteria, applying the same criteria uniformly to all inputs. 
Such static approaches cannot capture diverse perceptual aspects that vary across individuals. 
As illustrated in Figure~\ref{fig:intro}, different users attend to different aspects for the same image pair, leading to contrasting judgments--indicating that fixed, global dimensions are inadequate for personalized evaluation.

On the other hand, prior personalized T2I works~\cite{shen2024pmg, wang2024g, xu2025pigeon} commonly employ statistical metrics such as CLIPScore~\cite{hessel2021clipscore}, Fr\'echet Inception Distance (FID)~\cite{heusel2017gans}, and Inception Score (IS) ~\cite{salimans2016improved} to measure similarity between user references and generated outputs. 
However, such metrics capture only surface-level recurrence, oversimplifying the intricate nature of human aesthetic preferences underlying in personal history.
Some methods instead naïvely instruct LVLMs to assess preferences given reference data, but this lacks the in-depth reasoning to model latent preferences.
Assessing visual preference requires more than low-level similarity, also accounting for high-level semantic context.
Or they conduct human evaluations, which are inherently costly, time-consuming, and difficult to reproduce at scale.
These underscore the need for a more effective framework for assessing user-specific alignment in T2I generation.



In this work, we propose \textbf{\proposed}, a LVLM-based personalized reward model that (1) dynamically generates evaluation dimensions from user contexts, and (2) performs visual chain-of-thought (CoT) reasoning to score along those dimensions with explicit rationales.
To support such structured reasoning, we construct a 5K CoT-formatted dataset to train \proposed to learn to adaptively derive dimensions and perform context-aware evaluation across generated criteria, enabling interpretable assessment that captures diverse and complex user preferences.
Additionally, since real-world personalization often suffers from scarce user-level data~\cite{ryan2025synthesizeme}, we introduce a self-bootstrapping strategy~\cite{ye2025learning} that reasons over limited reference data to build rich user contexts.
These contexts allow \proposed\ to infer user-conditioned dimensions, enabling personalized assessment without user-specific training.

Further, we demonstrate the potential of \proposed\ beyond evaluation by leveraging its feedback for personalized prompt optimization.
Unlike prior works~\cite{li2024promptist, datta2024prompt} that aim to optimize prompts toward general aesthetic quality, \proposed\ incorporates individual context and thus  with individual context and thereby provides user-conditioned signal that reflect individual visual taste.
These signals identify preferred–rejected prompt variants, which are then used to fine-tune the prompt model via direct preference optimization (DPO), enabling user-aligned generation.

To evaluate real-world personalization in T2I generation, we additionally construct \textbf{\bench}, a user-level image preference dataset.
Existing datasets either focus on subtle visual differences~\cite{kirstain2023pick, wu2023human, liang2024rich}, lack contrastive feedback~\cite{chen2024tailored}, or are constrained to limited visual domains~\cite{salehi2024viper}.
In contrast, \bench\ is built from open-ended, abstract prompts that allow broad visual interpretation. 
Each instance consists of four images generated from expanded variants of a base prompt, and participants are asked to rank these images according to their own preferences.
For realistic personalization settings, we vary the number of instances provided to each participant, resulting in dynamic reference sizes.
In total, \bench\ comprises {75} user records, offering a realistic tested to assess how effectively T2I models infer and adapt to diverse human preferences.

\section{Related Work}
\label{sec:relwork}


\smallsection{Visual Chain-of-Thought Reasoning}
LVLMs~\cite{zhang2021mme, li2024llava, bai2025qwen2, wang2024qwen2} have demonstrated strong performance across diverse multimodal reasoning tasks~\cite{xiong2025llava, xu2025pigeon, leehippo}.
As CoT reasoning has significantly improved the performance of LLMs through step-by-step reasoning~\cite{wei2022chain, yao2023tree, kim2024self,lee2025imagine}, recent studies~\cite{zhang2023multimodal, zhang2024improve, zhao2025cot} have extended CoT reasoning to the vision--language domain, showing benefits across diverse visual reasoning tasks, including mathematics~\cite{gao2023g} and robotic control~\cite{zhao2025cot}. 
However, a key challenge in applying CoT to vision--language models lies in the scarcity of high-quality, CoT-annotated datasets. 
To address this, recent works~\cite{zhang2024improve, wang2025unifiedcot, dong2025insight} build data generation pipeline to collect task-specific CoT-formatted datasets and distill VLMs~\cite{wang2024qwen2, bai2025qwen2} under supervision.
Meanwhile, other works have explored LVLMs as generative reward models~\cite{li2024vlfeedback, zhou2024calibrated}, leveraging explicit rationales~\cite{shao2024deepseekmath, wang2025unified, xiong2025llava} for scalable and interpretable evaluation.
Incorporating  CoT has been shown to further enhance the reliability and robustness of reward modeling~\cite{zhang2024improve, wang2025unifiedcot}.


\smallsection{Personalized Text-to-Image Generation}
Unlike generic image generation, personalized T2I adapts outputs to user preferences by modeling intent from personal history (e.g., image interactions)~\cite{shen2024pmg, wang2024g,xu2025personalized}.
PIP~\cite{chen2024tailored} proposes a personalized prompt rewriting method conditioned on historical inputs. 
ViPer~\cite{salehi2024viper} and PPD~\cite{dang2025personalized} extract user features, expressed either as natural-language profiles or as hidden-state embeddings from LVLMs~\cite{li2024llava}, and directly inject these features into  diffusion models~\cite{pernias2023wurstchen}.
Recent methods expand to multi-turn settings by giving iterative feedbacks~\cite{nabatipreference, von2024fabric, zhang2025multi} or with natural language instructions through conversation~\cite{liu2024you, wang2024diffusion}.



\smallsection{Personalized Reward Modeling}
Preference alignment has been studied for large language models (LLMs), {particularly through} reinforcement learning from human feedback~\cite{stiennon2020learning}.
Subsequent works use pairwise preference data to align models with human interests~\cite{ouyang2022training, dubois2023alpacafarm, zhao2023slic, zheng2023secrets, seo-etal-2024-make}.
In particular, DPO~\cite{rafailov2023direct} has recently gained attention, also achieving great adaptation to diffusion models~\cite{fan2023dpok, su2024ddpo, wallace2024diffusion}.
However, recent studies highlight that human preferences are diverse~\cite{sorensen2024roadmap,chen2024pal}, motivating personalized reward modeling~\cite{rame2023rewarded, jang2023personalized}. 
For instance, GPG~\cite{zhang2024guided} and SynthesizeMe~\cite{ryan2025synthesizeme} introduce natural-language user profiles that encode personal context for conditioning; GPO~\cite{zhao2023group} targets group-level alignment by training an auxiliary predictor to steer models toward specific group preferences; VPL~\cite{poddar2024personalizing} formulates a latent-variable approach to capture multimodal, user-conditioned rewards; and PAL~\cite{chen2024pal} models heterogeneous preferences with a sample-efficient personalized reward framework.



\section{Personalized Reward Modeling \\ for Text-to-Image Generation}


\begin{figure}[t]
    \centering
    \includegraphics[width=0.99\linewidth]{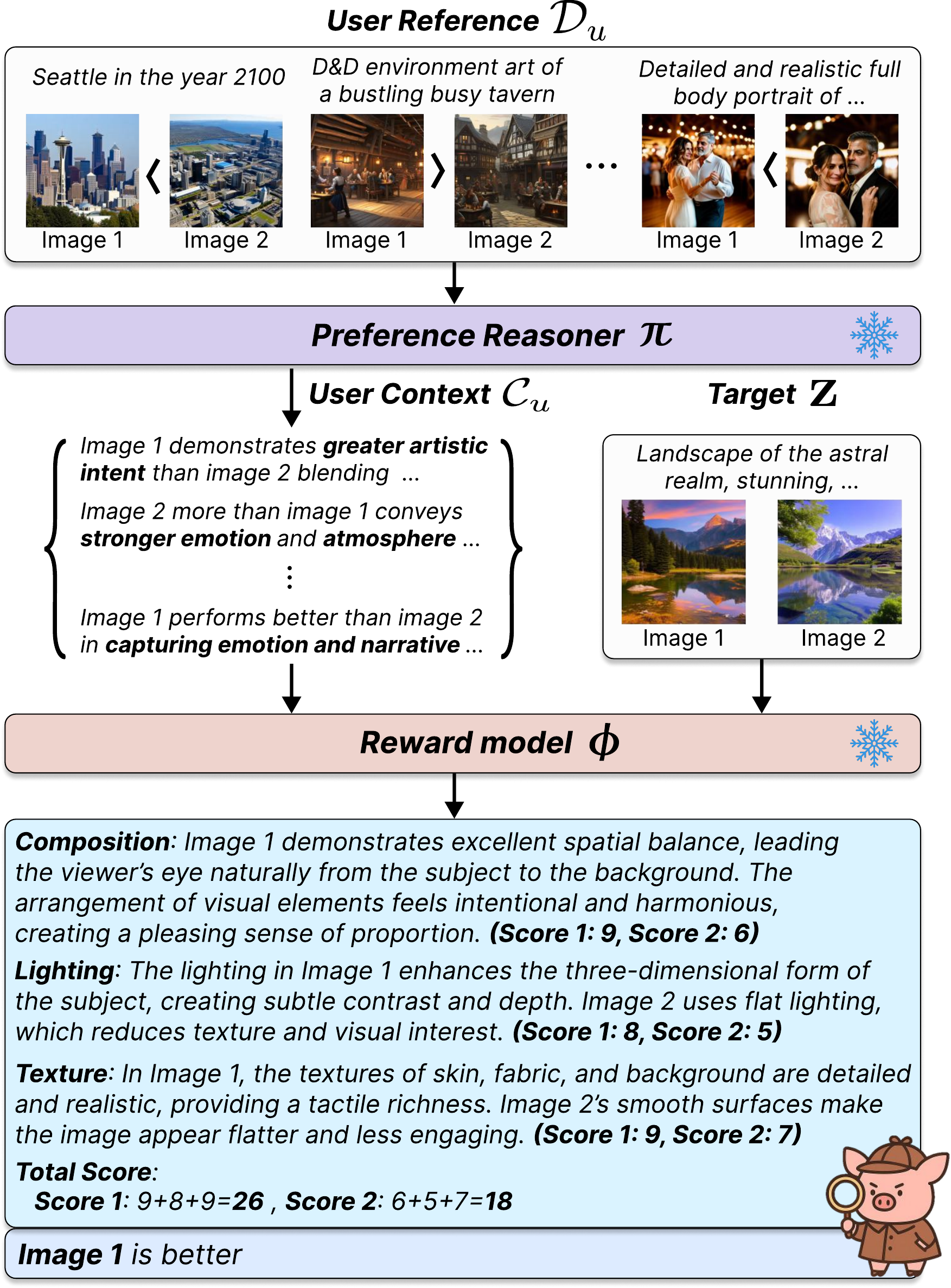}
    \caption{\textbf{Inference process in \proposed.} The preference reasoner $\pi$ first builds personalized context, then reward model $\phi$ performs context-aware evaluation of the target image pair.}
    \label{fig:inference}
    \vspace{-0.1cm}
\end{figure}


\smallsection{Problem Formulation}
The objective of personalized reward model $\mathcal{P}$ is to output a personalized reward $r$ for a target pair $\textbf{z} = (\mathbf{p}, \mathbf{I}_1, \mathbf{I}_2)$ given user reference data $\mathcal{D}_u = \{ (\mathbf{p}_i, \mathbf{I}_i^{+}, \mathbf{I}_i^{-}) \}_{i=1}^{N}$, where $\mathbf{p}_i$ is the prompt, $\mathbf{I}_i^{+}$ is the user-preferred image, and $\mathbf{I}_i^{-}$ is the non-preferred one:
\begin{equation}
\textbf{r} = \mathcal{P}(\textbf{z}, \mathcal{D}_u).
\end{equation}
Here, $\textbf{r}=+1$ if $\mathbf{I}_1$ is preferred, and $\textbf{r}=-1$ otherwise.

\subsection{\proposed}

We propose \proposed, that (1) models user preferences from reference data and (2) assesses two candidate images through generated evaluation dimensions.
The key idea of \proposed\ is to derive user-tailored dimensions by reasoning over limited reference data, facilitating personalized assessment without additional user-specific training.

\smallsection{Architecture Design}
The overview of \proposed is illustrated in Figure~\ref{fig:inference}.
We introduce the \emph{preference reasoner} $\pi$ which first processes $\mathcal{D}_u$ to construct 
a user context $\mathcal{C}_u$:
\begin{equation}
\mathcal{C}_u = \pi(\mathcal{D}_u) 
= \left\{ \mathbf{c}_i := \pi(\mathbf{p}_i, \mathbf{I}_i^{+}, \mathbf{I}_i^{-}) \right\}_{i=1}^{N}.
\label{eq:bootstrap}
\end{equation}
Conditioned on context $\mathcal{C}_u$, the \emph{reward model} $\phi$ evaluates the 
target pair and outputs the personalized reward $\mathbf{r}$:
\begin{equation}
\textbf{r} = \phi(\textbf{z}, \mathcal{C}_u).
\end{equation}
We adopt Qwen2-VL-7B~\cite{wang2024qwen2} as the backbone LVLM for both $\pi$ and $\phi$, though our framework is compatible with any LVLM capable of multi-image and language processing.
While we leverage the inherent multimodal reasoning capability of LVLMs, we fine-tune both $\pi$ and $\phi$ for task-specific objectives, as detailed below. 

\begin{figure}[t]
    \centering
    \includegraphics[width=0.99\linewidth]{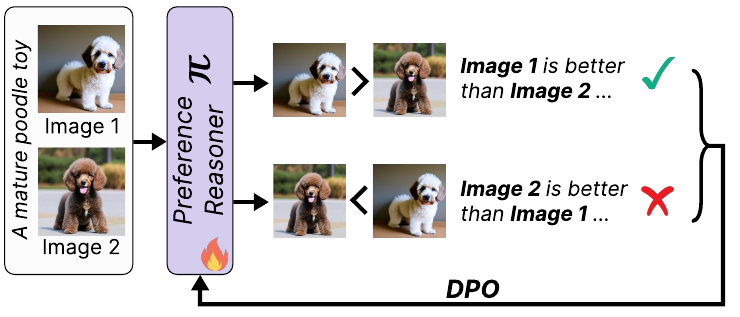}
    \vspace{-0.2cm}
    \caption{\textbf{Training process for the preference reasoner $\pi$.} We fine-tune $\pi$ with DPO by generating contrastive rationale pairs, enabling to produce multifaceted explanations for user preferences.}    
    \vspace{-0.2cm}
    \label{fig:reasoner_training}
\end{figure}

\smallsection{Fine-Tuning Preference Reasoner ${\pi}$}
To elicit preferences from few reference data, we bootstrap $\mathcal{D}_u$ using $\pi$ to generate rationales explaining why $\mathbf{I}_i^{+}$ is preferred over $\mathbf{I}_i^{-}$.
This process converts implicit image-level preferences into explicit, language-based reasoning.
However, vanilla LVLMs often lack the granularity and diversity required to articulate subtle visual preferences~\cite{trivedi2024self,ye2025learning,shypula2025evaluating}.  
To enhance their explanatory richness and lexical diversity, we train $\pi$ using contrastive judgments derived from a general preference dataset $\mathcal{D}_\mathcal{G} = \{ (\mathbf{p}_i, \mathbf{I}_i^{+}, \mathbf{I}_i^{-}) \}_{i=1}^{M}$.
For each triplet, we apply hint-driven sampling~\cite{ye2025learning}, where we instruct  $\pi$ to generate a {correct rationale} $j_i^{+}$ (explaining why $\mathbf{I}_i^{+}$ is preferred over $\mathbf{I}_i^{-}$), and an {incorrect rationale} $j_i^{-}$ (explaining why $\mathbf{I}_i^{-}$ is preferred over $\mathbf{I}_i^{+}$).  
We then optimize the preference reasoner $\pi$ via the DPO loss, where $x_i \coloneqq (\mathbf{p}_i, \mathbf{I}_i^{+}, \mathbf{I}_i^{-})$, and $\pi_{\text{ref}}$ is a frozen reference model (see Figure~\ref{fig:reasoner_training}).
\begin{equation}
\begin{split}
\mathcal{L}_{\pi}
&= -\,\mathbb{E}_{x_i,\, j_i^{+},\, j_i^{-}}
\Big[
    \log \sigma \big(\beta\, \Delta_i \big)
\Big], 
\text{where} \\[6pt]
\Delta_i 
&= 
\left(
    \log 
    \frac{
        \pi(j_i^{+}\!\mid x_i)
    }{
        \pi(j_i^{-}\!\mid x_i)
    }
    -
    \log 
    \frac{
        \pi_{\mathrm{ref}}(j_i^{+}\!\mid x_i)
    }{
        \pi_{\mathrm{ref}}(j_i^{-}\!\mid x_i)
    }
\right).
\raisetag{50pt}
\end{split}
\end{equation}
Here, $\beta>0$ is a temperature that controls the sharpness of the preference margin, and $\sigma(\cdot)$ denotes the logistic sigmoid that maps the scaled margin $\beta\, \Delta_i$ into a probability.

This objective encourages $\pi$  to assign higher relative likelihoods to correct rationales $j_i^{+}$ over incorrect ones $j_i^{-}$ for the same image pair, thereby improving the fidelity and diversity of generated explanations during bootstrapping.

\begin{figure}[t]
    \centering
    \includegraphics[width=0.99\linewidth]{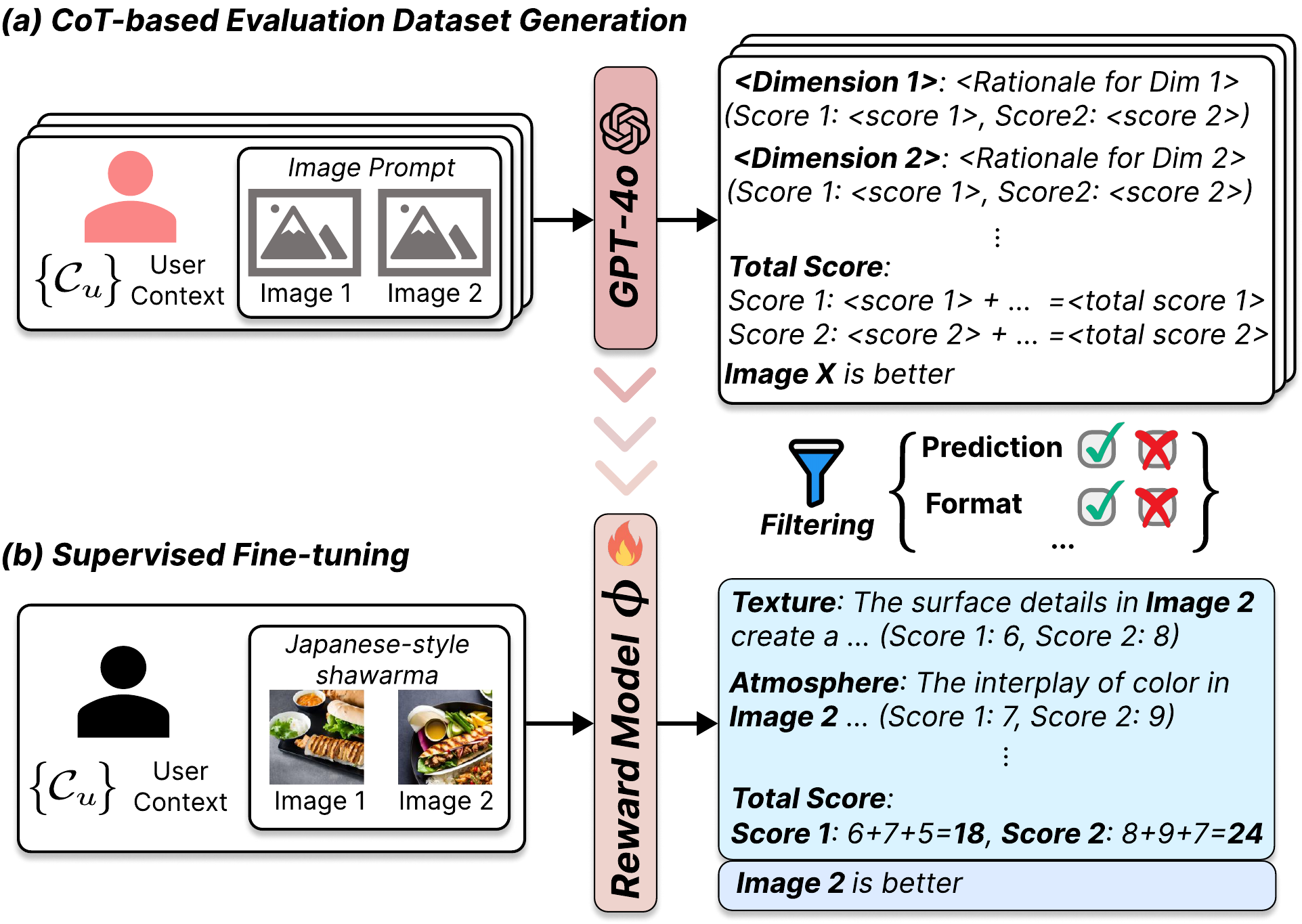}
    \caption{\textbf{Training process for the reward model $\phi$.}
    (a) We generate CoT-formatted data with GPT-4o and discard samples with any invalid or incorrect ones.
    (b) The resulting high-quality dataset is used to supervise $\phi$, distilling personalized reasoning process.}
    \vspace{-0.2cm}
    \label{fig:evaluator_training}
\end{figure}

\smallsection{Fine-tuning Personalized Reward Model $\phi$}
Given user context $C_u$, the reward model $\phi$ generates personalized evaluation for target image pair $x$ through detailed CoT-based reasoning which involves two key steps: 
(1) inferring user-specific evaluation dimensions and 
(2) assessing images accordingly into a final preference prediction. 

As shown in Figure~\ref{fig:evaluator_training}, we construct a dataset of 4K CoT-formatted samples to enable $\phi$ to learn structured and personalized reasoning.
To generate this data, we first prepare pairs of user context $\mathcal{C}_u$ and target $\textbf{z} = (\mathbf{p}, \mathbf{I}^+, \mathbf{I}^-)$.
The user context is written in natural language, containing rationales explaining the user’s previous preference decisions.
We then use GPT-4o~\cite{hurst2024gpt} to infer evaluation dimensions from user context and target images, then produce a detailed CoT reasoning that 
(i) scores each image along the inferred dimensions,
(ii) aggregates per-dimension scores, and 
(ii) identifies the preferred image. 
We filter out samples with inaccurate preference prediction or invalid format (e.g., fewer than three dimensions or incorrect score aggregation).
Then we supervise fine-tune $\phi$ on the resulting dataset as:
\begin{equation}
\mathcal{L}_{\phi} =
-\mathbb{E}_{(x_i, y_i)} 
\sum_{t=1}^{T_i}
\log \phi \big( y_{i,t} \mid {x}_i, y_{i,<t} \big),
\end{equation}
where ${x}_i = (\mathcal{C}_u, \textbf{z})$ includes both the user context and the target image pair, and $y_i = (y_{i,1}, \dots, y_{i,T_i})$ is the corresponding GPT-4o–generated CoT output.  
Through this training, $\phi$ learns to produce structured, dimension-aligned evaluations grounded in contextual user information.

\smallsection{Training Setup}
The data used to train $\pi$, train $\phi$, and evaluate are sampled from Pick-a-Pic~\cite{kirstain2023pick}, and each split is strictly disjoint.




\subsection{Personalized Prompt Optimization \\ with~\proposed}
\begin{figure}[t]
    \centering
    \includegraphics[width=0.99\linewidth]{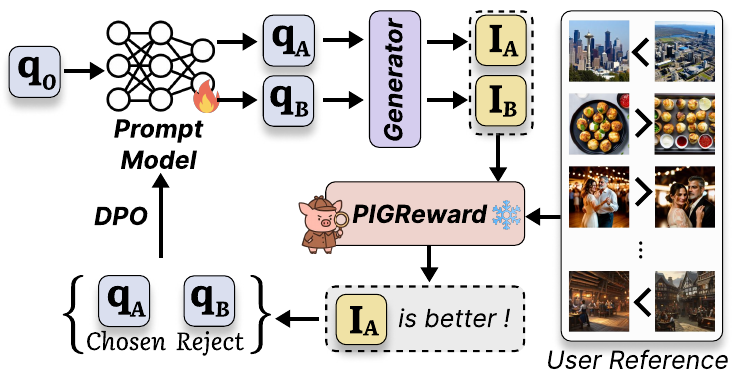}
    \caption{\textbf{An overview of personalized prompt optimization framework.} Here, \proposed\ offers user-conditioned guidance to refine the prompt model by identifying preferred–rejected pairs.}
    \label{fig:optimization}
\end{figure}

Beyond evaluation, \proposed can also serve as a personalized reward function for optimizing text prompts in T2I generation, guiding the prompt model toward generating images aligned with individual visual preferences. 
The overview is illustrated in Figure~\ref{fig:optimization}.

We train a prompt language model $\mathcal{M}$~\cite{li2024promptist} which refines initial prompt $\mathbf{q}_0$ into expanded variants.
For each user $u$, \proposed\ is instantiated as a personalized reward function $\mathcal{P}_u$ conditioned on reference data $\mathcal{D}_u$.
Given prompt $\mathbf{q}$, $\mathcal{M}$ generates two candidate expanded prompts, $\mathbf{q}_a$ and $\mathbf{q}_b$.
With a T2I generator $\mathcal{G}$, each candidate prompt is rendered into an image, as $\mathbf{I}_a = \mathcal{G}(\mathbf{q}_a)$ and $\mathbf{I}_b = \mathcal{G}(\mathbf{q}_b)$. 
Then $\mathcal{P}_u$ assesses two images then determines which image better aligns with the user preferences, resulting in chosen and rejected prompts:
\begin{equation}
\label{eq:prompt-choice}
(\mathbf{q}_i^{\mathrm{ch}}, \mathbf{q}_i^{\mathrm{rej}}) =
\begin{cases}
(\mathbf{q}_a, \mathbf{q}_b), & \text{if } \mathcal{P}_u(\mathbf{I}_a, \mathbf{I}_b) = +1,\\[3pt]
(\mathbf{q}_b, \mathbf{q}_a), & \text{otherwise.}
\end{cases}
\end{equation}

The identified chosen-reject pairs form user-conditioned preference signal for DPO training of $\mathcal{M}$.
Let $x_i := (\mathbf{q}_{0,i}, u)$ denote the conditioning context, abstract prompt and user feature. 
We then optimize $\mathcal{M}$ using the DPO objective:
\begin{equation}
\begin{split}
\mathcal{L}_{\mathcal{M}}
&= -\mathbb{E}_{(x_i,\,\mathbf{q}_i^{\mathrm{ch}},\,\mathbf{q}_i^{\mathrm{rej}})}
\left[\log \sigma(\beta \Delta_i)\right], \\
\text{where}\quad
\Delta_i
&= \log \frac{\mathcal{M}(\mathbf{q}_i^{\mathrm{ch}}\mid x_i)}{\mathcal{M}(\mathbf{q}_i^{\mathrm{rej}}\mid x_i)}
     - \log \frac{\mathcal{M}_{\mathrm{ref}}(\mathbf{q}_i^{\mathrm{ch}}\mid x_i)}{\mathcal{M}_{\mathrm{ref}}(\mathbf{q}_i^{\mathrm{rej}}\mid x_i)}.
\raisetag{45pt}
\end{split}
\end{equation}
This training encourages $\mathcal{M}$ to increase the likelihood of prompts that yield higher personalized rewards from \proposed, producing a user-adaptive prompt generator that aligns more closely with individual visual preferences.




\begin{figure}[t]
    \centering
    \includegraphics[width=\linewidth]{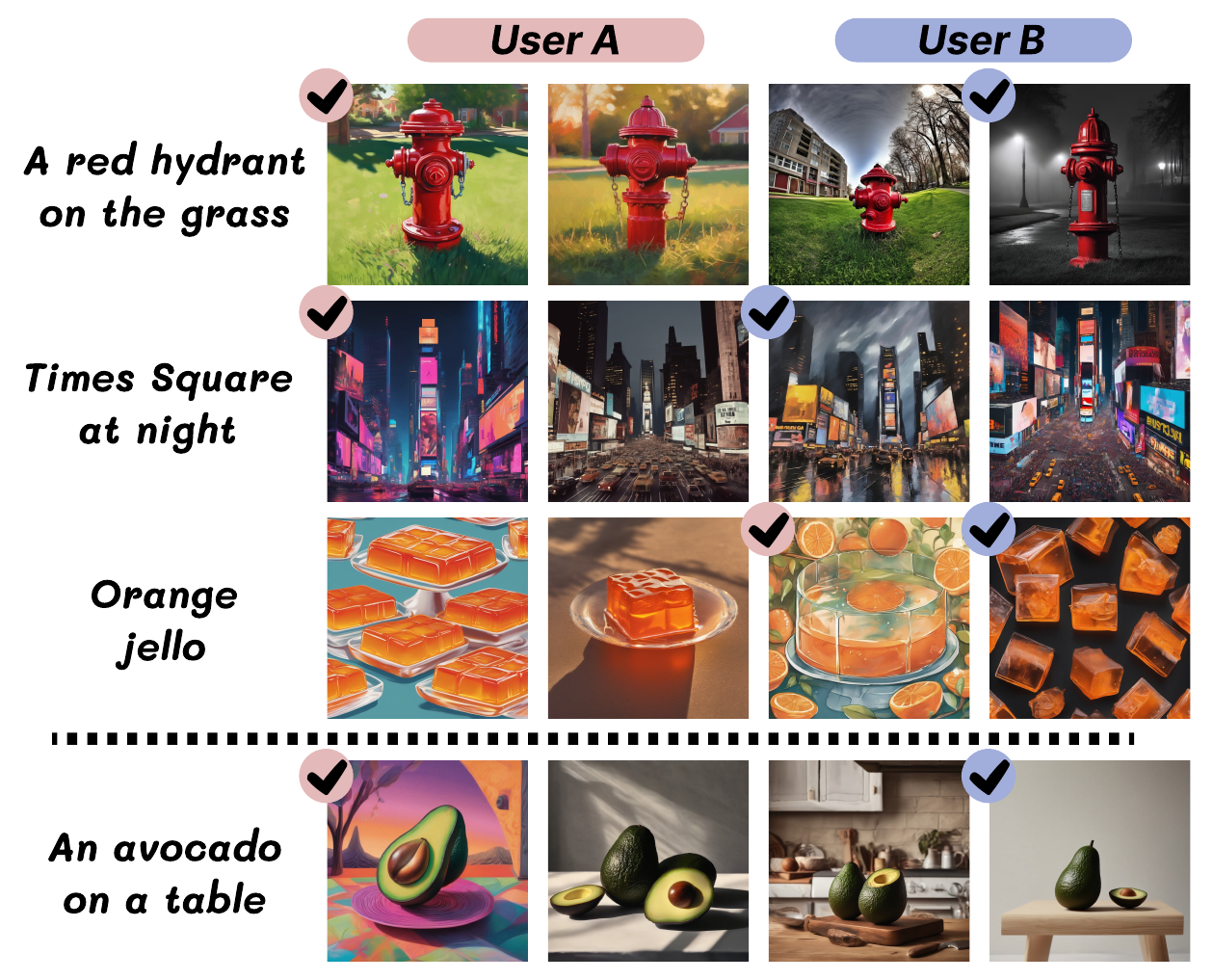}
    \caption{
    \textbf{Examples from \bench.} Each instance contains four image variants generated from the same prompt. The checkmarked image indicates the choice of User A or B. Users show distinct visual preferences across instances (above the dashed line) and even select different images within the same instance (below the dashed line), reflecting divergent aesthetic tastes.
    }
    \label{fig:pigbench}
\end{figure}

\section{PIGBench: A Personalized T2I Dataset}

\smallsection{Discussing Existing Datasets}
Although several user-level preference datasets exist, they pose challenges for evaluating personalization.
Pick-a-Pic~\cite{kirstain2023pick} is the most widely used pairwise preference dataset and includes user identifiers, allowing user-level analysis.
However, as it was originally designed to capture general human preferences rather than individual characteristics, it lacks distinct user-specific variation~\cite{dang2025personalized}.
PIP~\cite{chen2024tailored} is the first T2I dataset explicitly aimed at personalization, consisting of user-written prompts and generated images. 
Yet it lacks contrastive information--what users dislike--overlooking the importance of negative feedback in modeling preference~\cite{mo2025learning}. 
ViPer~\cite{salehi2024viper} provides images with like/dislike comments simulated by LVLMs, but its data is limited to artistic domains, creating a gap from real-world imagery.

\smallsection{Introducing \bench}
We therefore construct \bench, a benchmark dataset to evaluate real-world personalization in T2I generation.
Unlike prior dataset, \bench focuses on how simple textual prompts can be visually interpreted according to diverse user preferences. 
\bench comprises user-level instances, where each instance includes a user-specific ranking of four images generated from a single, abstract prompt. 
To reflect realistic settings, we vary the amount of reference data per user.
This allows us to assess not only how well T2I models align with individual user preferences, but also how effectively they model preferences from dynamic context lengths.


\smallsection{Constructing Dataset}
We take the ``\textit{simple detail}'' subset of PartiPrompts~\cite{yu2022scaling}, which leaves substantial room for specification, and curate prompts spanning diverse 12 categories (e.g., world knowledge, outdoor scenes).
For each prompt, we use GPT-4o~\cite{hurst2024gpt} to expand it into four concrete variants that differ in visual style and semantic context, then generate the corresponding images with SDXL~\cite{podell2023sdxl}.
We then conduct a manual quality check to remove low-quality sets and  filter out sets with insufficient personalization affordance, where the variants are unlikely to elicit divergent user preferences, resulting in 100 instances in total.
Human annotators are asked to rank the four images for each expanded prompt according to their personal preferences. 
Each annotator is randomly assigned a varying number of image sets (between 5 and 15) to collect records with various length of user context, there by resulting in 75 user records.
Figure~\ref{fig:pigbench} shows examples from \bench, highlighting that different users often show distinct rankings for the same image sets.

\begin{table}[t]
\centering
\caption{\textbf{Dataset statistics.} Reference size denotes the number of samples per user reference, which varies across datasets.}
\vspace{-0.2cm}
\resizebox{\linewidth}{!}{%
\begin{tabular}{lcccc}
\toprule
 & \textbf{Pick-a-Pic} & \textbf{PIP} & \textbf{PASTA} & \textbf{PIGBench} \\
\midrule
Reference size & 8 & 5 & 4 & 5-15 \\
Users & 10,000 & 1,002 & 5,143 & 75 \\
\bottomrule
\end{tabular}%
}
\vspace{-0.2cm}

\label{tab:dataset_stats}
\end{table}

\begin{table*}[t]
\centering
\caption{\textbf{Quantitative comparison with conventional T2I evaluation methods.} Baselines are evaluated without user references, and we compare against vanilla LVLMs as well given references. The best results are in bold, and the second-best results are \underline{underlined}.}
\resizebox{0.99\linewidth}{!}{
\begin{tabular}{l|c c|c c|c c|c c}
\toprule
 & \multicolumn{2}{c|}{\textbf{Pick-a-Pic}} 
 & \multicolumn{2}{c|}{\textbf{PIP}} 
 & \multicolumn{2}{c|}{\textbf{PASTA}} 
 & \multicolumn{2}{c}{\textbf{PIGBench}} \\
 & acc w/ tie & acc w/o tie 
 & acc w/ tie & acc w/o tie 
 & acc w/ tie & acc w/o tie 
 & acc w/ tie & acc w/o tie \\
\midrule
\rowcolor{gray!10}
\multicolumn{9}{c}{\textit{w/o Reference}} \\
CLIPScore~\cite{hessel2021clipscore}       & 50.90 & 55.44 & 49.15 & 21.37 & 66.28 & 71.09 & 47.17 & 26.42 \\
Aesthetics~\cite{schuhmann2022laion}                            & 49.98 & 25.43 & 51.66 & 37.82 & 51.78 & 59.56 & 49.06 & 26.42 \\
PickScore~\cite{kirstain2023pick}          & 51.00 & 54.31 & 49.63 & 33.87 & 68.36 & 71.13 & 51.89 & 16.98 \\
HPS (v2)~\cite{wu2023human}                               & 50.14 & 6.65  & 49.68 & 12.61 & 69.03 & \underline{72.99} & 43.40 & 28.30 \\
ImageReward~\cite{xu2023imagereward}       & 49.55 & 41.81 & 49.79 & 43.06 & 69.82 & 70.03 & 35.85 & 33.96 \\
VisionReward~\cite{xu2024visionreward}     & 49.81 & 30.40 & 49.73 & 31.73 & 65.54 & 71.18 & 43.40 & 34.78 \\
MPS (overall)~\cite{zhang2024learning}     & 50.20 & 47.87 & 50.80 & 46.15 & 66.73 & 68.17 & 47.17 & 45.28 \\
LLava-Reward~\cite{zhou2025multimodal}                   & 41.57 & 48.95 & 25.60 & 25.48 & 68.17 & 50.57 & 64.15 & 64.15 \\
UnifiedReward~\cite{wang2025unified}                              & 48.46 & 48.46 & 37.48 & 37.41 & \underline{69.86} & 69.86 & \underline{67.92} & 67.92 \\
UnifiedReward-Think~\cite{wang2025unifiedcot} & 59.92 & 59.92 & 35.28 & 35.21 & 65.71 & 65.71 & 66.04 & 66.04 \\
\midrule
\rowcolor{gray!10}
\multicolumn{9}{c}{\textit{w/ Reference}} \\
Qwen2-VL-7B~\cite{bai2025qwen2}            & \underline{61.52} & \underline{61.52} & 49.95 & 49.95 & 50.34 & 50.34 & 54.72 & 54.72 \\
LLaVA-OneVision~\cite{li2024llava}         & 55.04 & 55.04 & 50.75 & 50.76 & 50.12 & 50.12 & 56.60 & 56.60 \\
GPT-4o~\cite{hurst2024gpt}                & 46.90 & 46.70 & \underline{76.23} & \underline{76.54} & 68.80 & 68.80 & 64.15 & \underline{68.29} \\
\midrule
\textbf{\proposed}                         & \textbf{63.76} & \textbf{62.14} & \textbf{77.84} & \textbf{78.59} & \textbf{75.43} & \textbf{74.43} & \textbf{84.91} & \textbf{85.85} \\
\bottomrule
\end{tabular}
}
\label{tbl:main}
\end{table*}
\begin{figure*}[t]
\centering


\includegraphics[width=\linewidth]{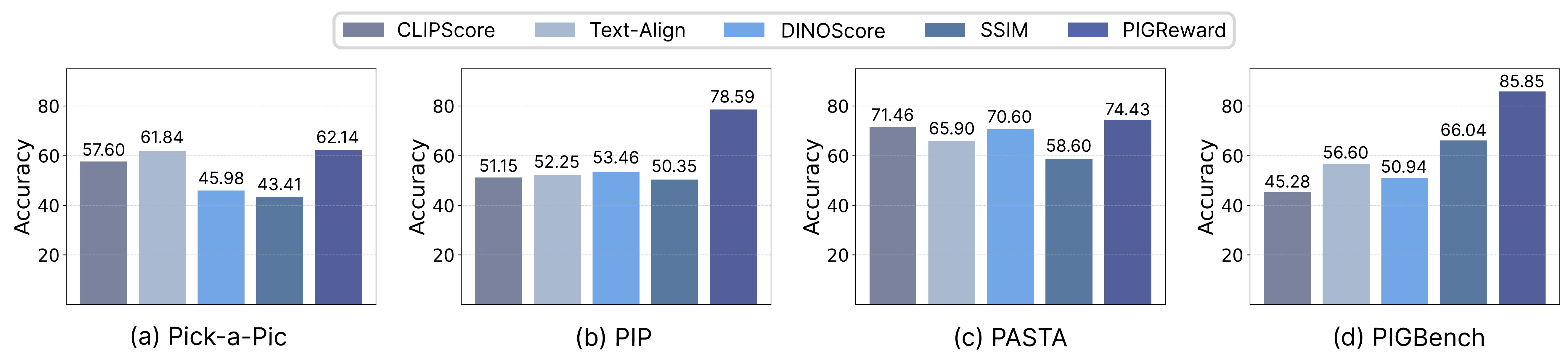}
\vspace{-0.6cm}
\caption{\textbf{Quantitative comparison with similarity-based metrics.} CLIPScore, DINOScore, and SSIM compare generated images to reference images, while Text-Align compares to reference prompts. PIGReward leverages both reference images and prompts.}
\label{fig:similarity_comp}
\vspace{-0.1cm}
\end{figure*}

\section{Experiments}
\subsection{Experimental Settings}
\smallsection{Datasets}
We evaluate on three preference datasets—Pick-a-Pic~\cite{kirstain2023pick}, PIP~\cite{chen2024tailored}, and PASTA~\cite{nabatipreference}—and \bench. 
Dataset statistics are in Table~\ref{tab:dataset_stats}.
Although they contain user identifiers, they are not intentionally designed for history-driven personalized evaluation, we adapt them to a user-conditioned pairwise protocol.



\smallsection{Baselines}
We compare \proposed\ against conventional and recent state-of-the-art T2I evaluation methods. 
(1) \textbf{w/o Reference:} 
Conventional evaluation methods assess image quality without user-specific context.  
We include CLIP/BLIP-based reward models such as CLIPScore~\cite{hessel2021clipscore}, PickScore~\cite{kirstain2023pick}, HPS~\cite{wu2023human}, Aesthetics~\cite{schuhmann2022laion}, and ImageReward~\cite{xu2023imagereward}, all of which are fine-tuned on general single-dimensional preference datasets.
We further consider multi-dimensional approaches such as MPS~\cite{zhang2024learning} and VisionReward~\cite{xu2024visionreward}, which include fixed criterias.  
Lastly, we include recent MLLM-based reward models: LLaVA-Reward~\cite{xiong2025llava} and UnifiedReward~\cite{wang2025unified}, as well as UnifiedReward-Think~\cite{wang2025unified}, which incorporates CoT reasoning process.
(2) \textbf{w/ Reference:}
Previous personalized T2I studies~\cite{dang2025personalized, xu2025personalized} adopt the LVLM-as-a-judge paradigm, in which a large vision-language model (LVLM) evaluates image pairs based on given reference examples.
We employ two open-source models—Qwen2-VL-7B and LLaVA-OneVision~\cite{li2024llava}—which support multi-image reasoning, and one closed-source model, GPT-4o~\cite{hurst2024gpt}.

\subsection{Main Results}
Table~\ref{tbl:main} presents the results of \proposed\ and existing T2I evaluation methods under a personalized setting, where \proposed\ consistently achieves the best performance.
Both single-scalar and fixed multi-dimensional baselines underperform, indicating that a limited set of evaluation axes is insufficient to capture fine-grained human preference and user-dependent preferences, highlighting the need for adaptive assessment.
Even MLLM-based method perform comparably to simple reward-function baselines, indicating that generic reasoning alone is insufficient to tailor individuals.
Among baselines without reference data, UnifiedReward-Think shows the strong performance, demonstrating that incorporating CoT reasoning significantly enhances evaluation capability.
When provided with reference data, vanilla LVLMs show improved results but still fall short of \proposed, suggesting that they struggle to infer complex and latent preferences embedded in user histories.
Overall, these results confirm that context-conditioned evaluation dimensions and detailed CoT-based reasoning are both essential for effective personalized evaluation.


\subsection{Comparisons with Similarity-based Metrics}
Previous personalized T2I works~\cite{chen2024tailored, kim2025draw} typically evaluate personalization with similarity-based metrics that measures how closely generated outputs resemble reference images or prompts.
We include CLIPScore (Text-Align), SSIM, and DINOScore in our comparison: CLIPScore and Text-Align compute CLIP-based similarities~\cite{hessel2021clipscore}, SSIM captures pixel-level perceptual similarity~\cite{wang2003multiscale}, and DINOScore measures visual similarity in the DINO feature space~\cite{oquab2023dinov2}.
As shown in Figure~\ref{fig:similarity_comp}, these metrics correlate weakly  with actual user preferences, as they are limited to capturing surface-level resemblance rather than contextual intent.
Pixel- or embedding-level similarity often overlook subjective visual preference aspects, resulting in low sensitivity to user-specific preferences, especially across diverse users or abstract prompts.
In contrast, \proposed\ reasons over user-specific context, enabling a more faithful understanding of individual preference than fixed similarity measures.


\begin{table}[t]
\centering
\caption{\textbf{Ablation on training stages and CoT reasoning.}
Evaluates the impact of CoT reasoning, DPO-tuned preference reasoner $\pi$, and CoT-distilled reward model $\phi$.}
\vspace{-0.2cm}
\resizebox{0.89\linewidth}{!}{%
\begin{tabular}{c|cccc}
\toprule
 & \textbf{DPO $\pi$} & \textbf{SFT $\phi$} & \textbf{Pick-a-Pic} & \textbf{PIGBench} \\
\midrule
\multirow{1}{*}{\textit{w/o CoT}} 
 & \cmark & \xmark & 44.43 & 39.62 \\
\midrule
\multirow{4}{*}{\textit{w/ CoT}} 
 & \xmark & \xmark & 48.94 & 56.94 \\
 & \xmark & \cmark & 49.98 & 58.49 \\
 & \cmark & \xmark & 45.63 & 70.83 \\
 & \cmark & \cmark & \textbf{63.76} & \textbf{84.91} \\
\bottomrule
\end{tabular}%
}
\label{tbl:train_ablation}
\end{table}

\begin{figure}[t]
    \centering
    \includegraphics[width=0.79\linewidth]{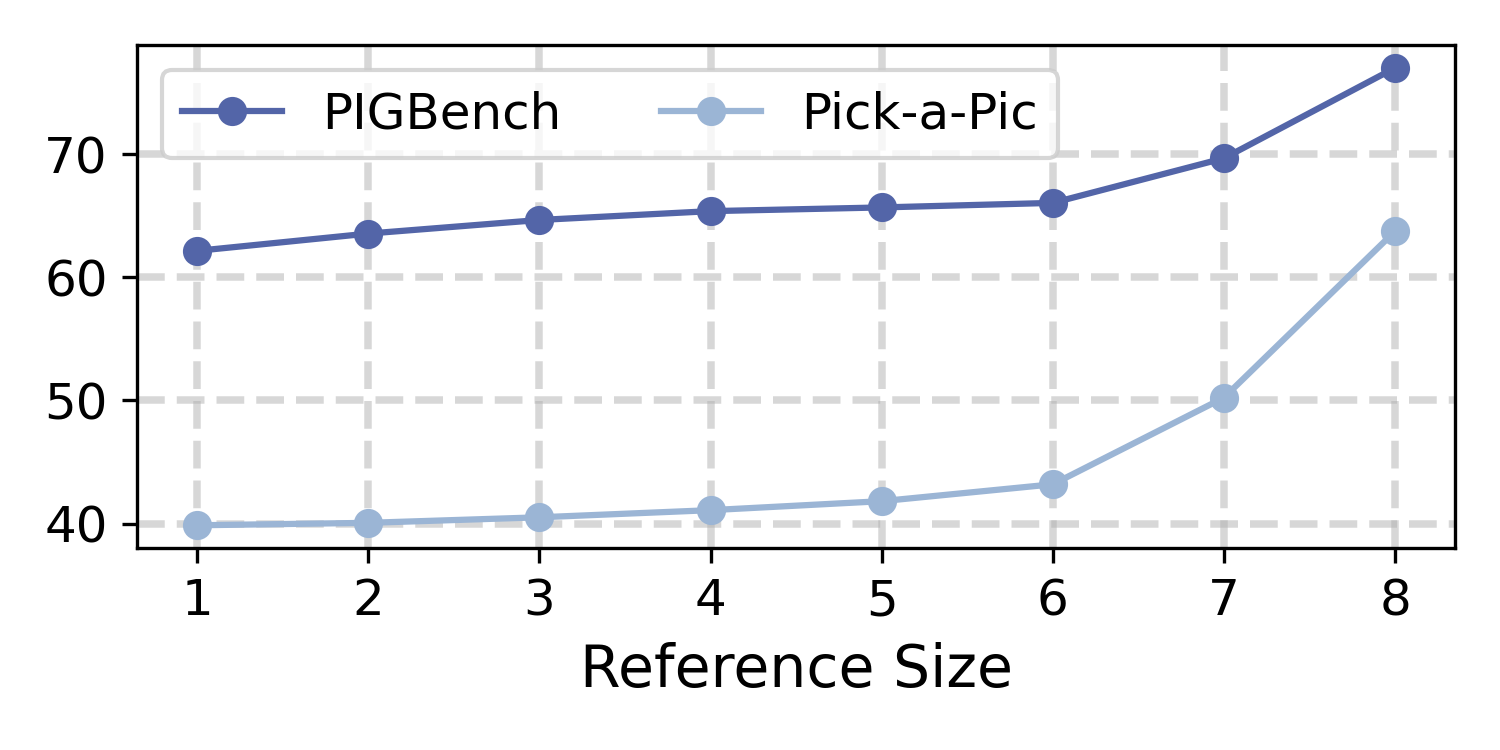}
    \vspace{-0.3cm}
    \caption{\textbf{Ablation of varying reference size.} We control the reference data size per user from 1 to 8.}
    \label{fig:history_length}
    \vspace{-0.4cm}
\end{figure}

\subsection{Ablation Studies}
\label{subsec:ablation}
\smallsection{Ablation of Each  Training Stage and CoT}
Table~\ref{tbl:train_ablation} reports the results of our ablations across training stages and the impact of CoT reasoning.
Without CoT reasoning, \proposed performs the worst, as it lacks a step-by-step, dimension-aware evaluation process.
Introducing CoT reasoning improves performance by enabling structured, user-specific reasoning.
Fine-tuning the preference reasoner $\pi$ with DPO further boosts performance by producing more diverse and discriminative rationales.
The best results are achieved when the reward model $\phi$ is further distilled on personalized CoT-formatted data, refining the ability of  \proposed\ to effectively perform personalized evaluation.

\smallsection{Ablation of User Context Length}
We study how the size of user reference affects personalization performance.
Figure~\ref{fig:history_length} shows that performance steadily improves as the size increases from 1 to 8. 
This result exhibits the importance of richer user context in accurately predicting individual preferences, as longer interaction histories enable \proposed to infer more fine-grained representations of user intent.

\begin{figure}[t]
    \centering
    \includegraphics[width=0.89\linewidth]{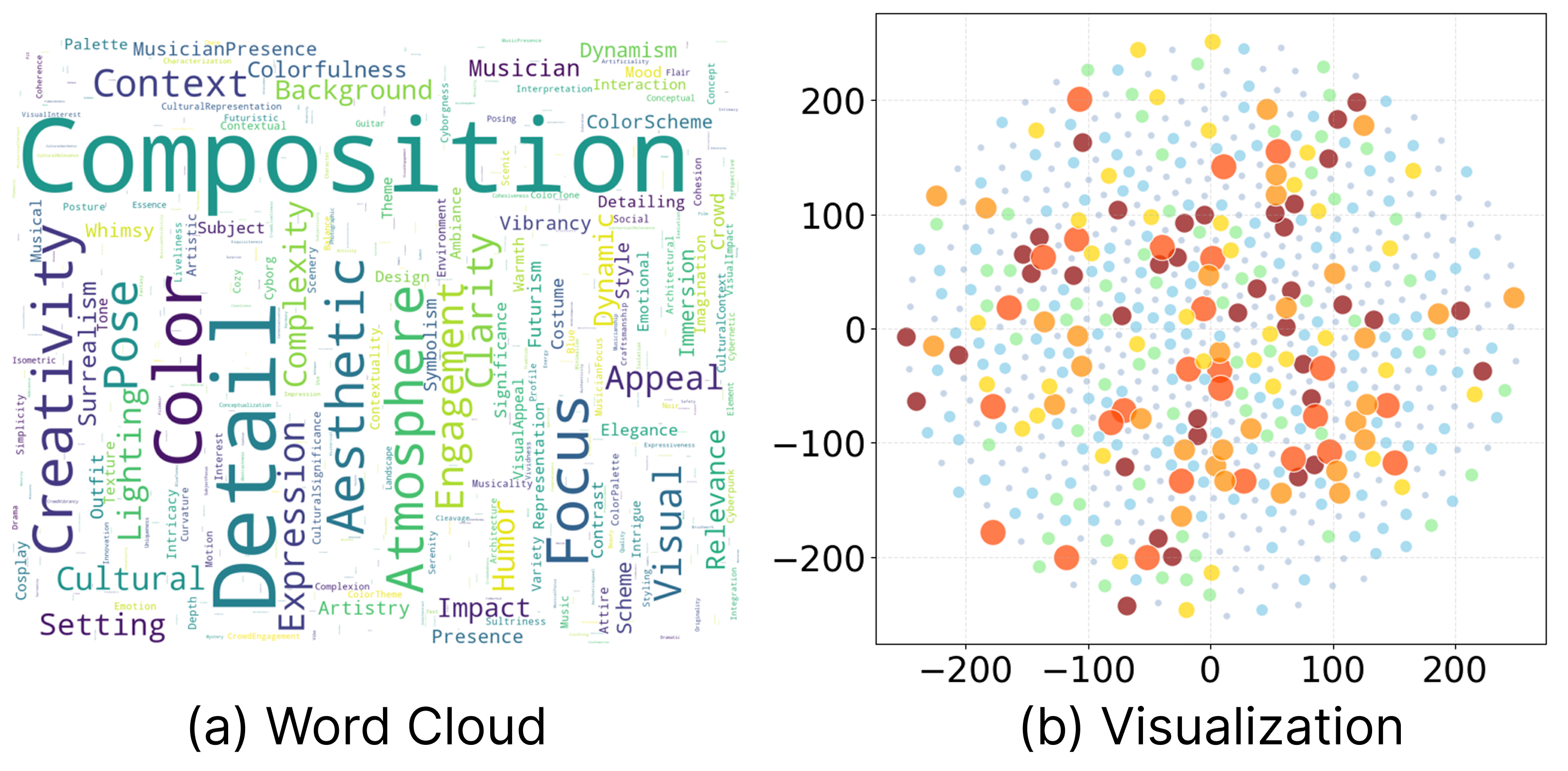}
    \vspace{-0.1cm}
    \caption{\textbf{Visualization of evaluation dimensions} generated by \proposed. In (b), larger dots indicate dimensions that appear more frequently, reflecting common evaluative tendencies.}
    \label{fig:dynamic_dimensions}
    \vspace{-0.2cm}
\end{figure}

\begin{figure}[t]
    \centering
    \includegraphics[width=0.99\linewidth]{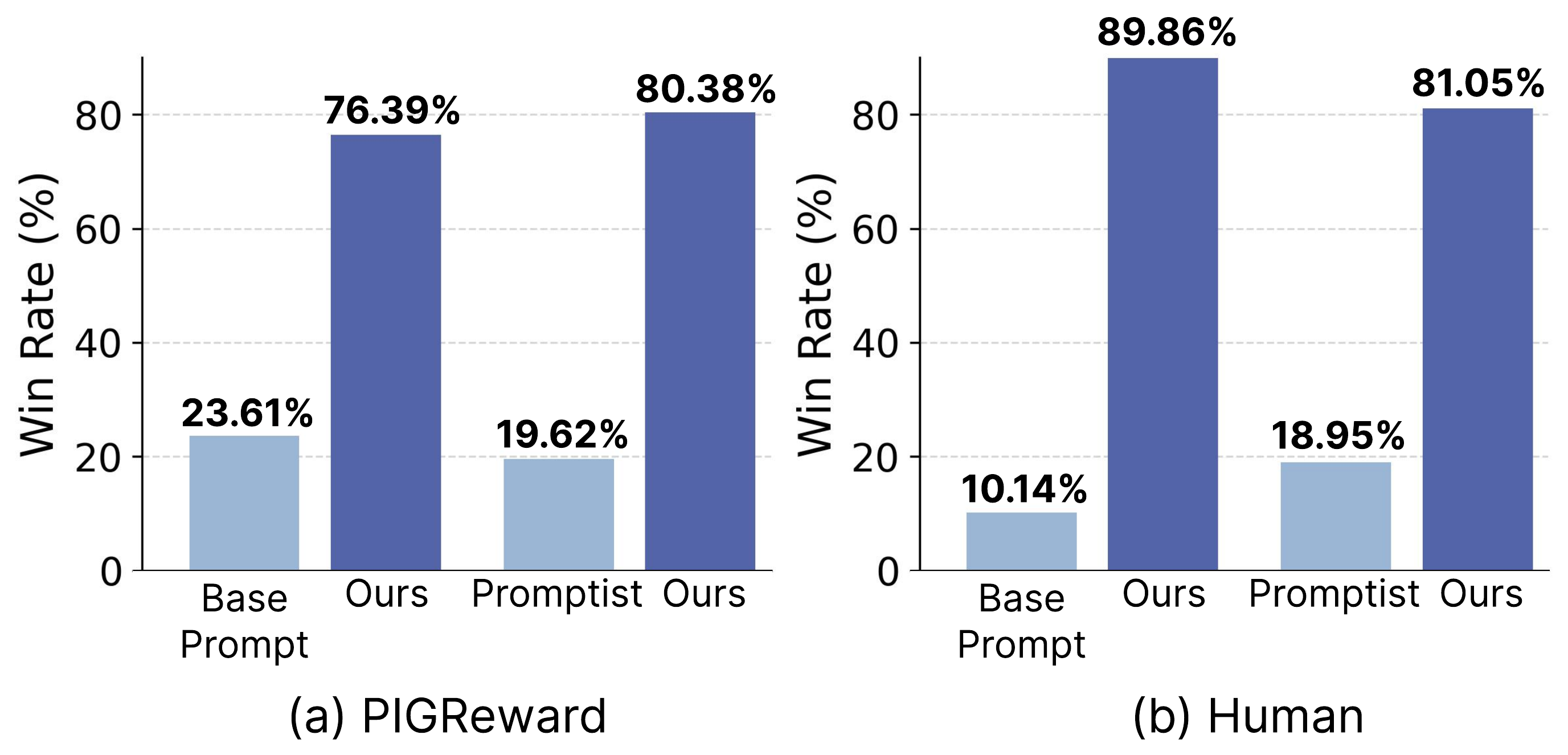}
    \vspace{-0.2cm}
    \caption{\textbf{Quantitative results for personalized prompt optimization.} Win rates are computed using (a) automatic evaluation by \proposed\ and (b) human evaluation.}
    \label{fig:optimized_winrate}
    \vspace{-0.3cm}
\end{figure}

\begin{figure*}[t]
    \centering
    \includegraphics[width=0.99\linewidth]{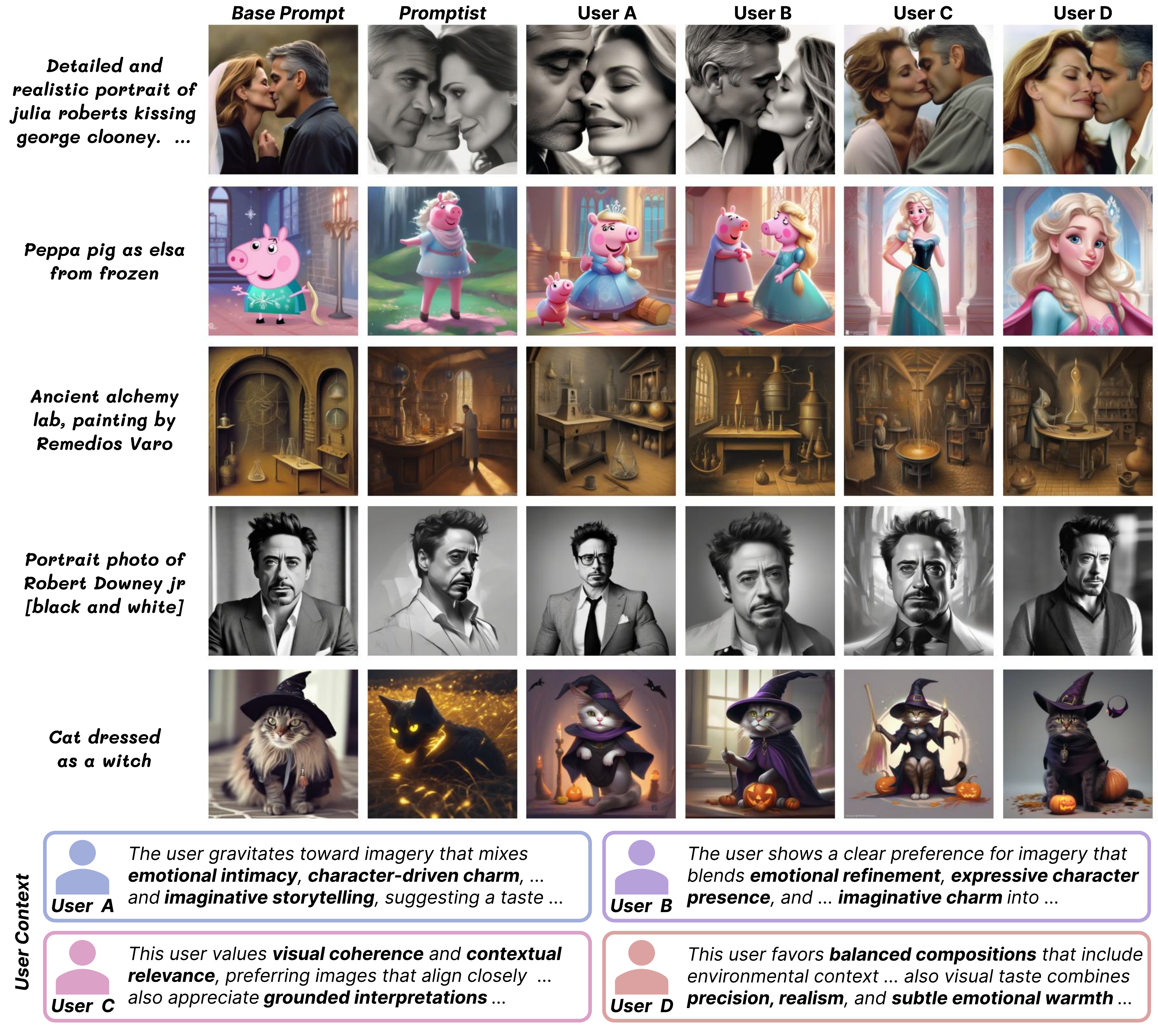}
    \vspace{-0.2cm}
    \caption{\textbf{Qualitative results for personalized prompt optimization.} 
    We show images generated from base prompts, Promptist-refined prompts, and prompts optimized by \proposed\ under four distinct user contexts, each reflecting different visual preferences. The depicted user contexts are derived from our preference reasoner and summarized using an LLM. For example, User A favors imaginative narrative elements, leading to more illustrative outputs, whereas User D prioritizes realism, resulting in highly detailed images with natural lighting.}
    \label{fig:optimized}
    \vspace{-0.2cm}
\end{figure*}

\subsection{Analysis of Generated Evaluation Dimensions}
The core idea of \proposed is to dynamically derive user-conditioned evaluation dimensions from personalized contexts, unlike prior approaches that rely on predefined criteria.
We visualize the generated dimensions during evaluation in Figure~\ref{fig:dynamic_dimensions}: (a) a word cloud of frequent terms and (b) t-SNE embeddings.
\proposed\ generates a diverse set of descriptive terms with well-dispersed embeddings, indicating that the inferred dimensions capture distinct and varied evaluation perspectives.
We can conclude that fixed criteria cannot capture the richness of user-dependent factors.

\subsection{Quantitative and Qualitative Study for Personalized Prompt Optimization}
We compare images generated from prompts optimized by \proposed\ with those produced from the base prompts and from prompts optimized by Promptist~\cite{li2024promptist}, a general-purpose method.
We use SDXL~\cite{podell2023sdxl} for generator.
To initialize \proposed, we use user contexts constructed from the \bench\ dataset, and the same annotators who created \bench also conducted the quantative evaluation.
Additionally, we include automatic evaluation using \proposed\ itself.
As shown in Figure~\ref{fig:optimized_winrate}, ours achieve a substantially higher win rate, indicating that feedback from \proposed\ effectively guides the generation process toward user-aligned outputs.
Figure~\ref{fig:optimized} presents qualitative examples of personalized prompt optimization.
Across different users, the optimized prompts lead to visually diverse outputs, demonstrating that effective personalization requires tailoring the optimization process to individual preferences rather than following a single, globally optimal direction.

\section{Conclusion and Future Directions}
In this work, we introduced \proposed, a personalized reward model for evaluating and optimizing T2I generation.
\proposed\ infers user-specific evaluation dimensions and performs multi-dimensional CoT reasoning, enabling fine-grained, context-aware assessments.
We also show its effectiveness for personalized prompt optimization, where user-conditioned rewards guide generation toward individual visual preferences.
To support realistic evaluation, we present \bench, a benchmark capturing diverse user preferences from abstract prompts.
Experiments show that \proposed\ surpasses existing methods in accuracy and interpretability, establishing a strong foundation for personalized T2I evaluation and optimization.
Future works can leverage the reasoning traces generated by \proposed\ as natural language feedback to guide T2I generation models.
Since these explanations provides the detailed rationale for preference 
prediction, highlighting why certain images succeed or fail to satisfy user-specific attributes, they can serve as informative guidance for 
prompt refinement and model alignment.
This direction opens the door to more interactive, interpretable, and individual-aligned generative systems.

{
    \small
    \bibliographystyle{ieeenat_fullname}
    \bibliography{main}
}


\end{document}